\documentclass[11pt,a4paper]{erlarticle}
\usepackage{mathtools}
\usepackage{graphicx}
\usepackage{booktabs}
\usepackage{array}
\usepackage{tabularx}
\usepackage{placeins}
\usepackage{needspace}
\usepackage{float}
\definecolor{frozenblue}{RGB}{232,240,249}
\definecolor{gafred}{RGB}{252,232,226}
\definecolor{perturbgray}{RGB}{242,244,246}
\definecolor{lightingyellow}{RGB}{255,248,224}
\definecolor{distractorpurple}{RGB}{244,238,249}

\newcommand{\smolvla}{SmolVLA}
\makeatletter
\newcommand{\bstctlcite}[1]{\@bsphack\if@filesw\immediate\write\@auxout{\string\citation{#1}}\fi\@esphack}
\makeatother
\title{Guided Action Flow: Value-Guided Sampling for\\Frozen Vision-Language-Action Policies}
\runningtitle{Guided Action Flow}
\author[1]{Liuhaichen~Yang\textsuperscript{\textdagger}}
\author[1]{Zhuang~Jiang\textsuperscript{\textdagger}}
\author[1]{Chenchao~Sheng\textsuperscript{\textdagger}}
\author[1]{Ningwei~Bai}
\author[1]{Qichen~Yin}
\author[2]{Hanbo~Ma}
\author[3]{Junkai~Liu}
\author[3]{Junkai~Sun}
\author[3]{Dongcheng~Lyu}
\author[4]{Yi~Dong}
\author[1]{Zezhi~Tang\textsuperscript{*}}
\newcommand{\gafaffiliations}{%
  \mbox{\textsuperscript{1}University College London}\quad
  \mbox{\textsuperscript{2}The University of Manchester}\\
  \mbox{\textsuperscript{3}Nanjing Institute of Software Technology, Chinese Academy of Sciences}\quad
  \mbox{\textsuperscript{4}University of Liverpool}}
\affil[ ]{\protect\gafaffiliations}
\titlenote{\textsuperscript{\textdagger}Equal contribution.\quad
  \textsuperscript{*}Corresponding author: Zezhi Tang. E-mail: \href{mailto:zezhi.tang@ucl.ac.uk}{\textcolor{ERLMuted}{\texttt{zezhi.tang@ucl.ac.uk}}}.}
\date{}
\begin{document}
\bstctlcite{GAFbibcontrol}
\maketitle

\begin{abstract}
Reinforcement learning can improve vision-language-action (VLA) policies beyond supervised fine-tuning, although this typically involves further updates to the policy parameters. For flow-matching policies, iterative action generation provides an additional opportunity to incorporate task information during inference. We introduce Guided Action Flow (GAF), which learns a compact, observation-conditioned action-value critic from robot task rollouts and applies its action gradient to steer reverse-time flow sampling. The supervised-fine-tuned VLA remains frozen throughout critic learning and deployment. Physical-robot experiments show an increase in aggregate success from 60.0\% to 82.5\% across six nominal manipulation tasks. Under six altered-lighting and object-distractor conditions evaluated on three of these tasks, aggregate success improves from 34.2\% to 49.2\%. Ablations and rollout analyses support the importance of the learned guidance direction and the critic's visual and proprioceptive inputs. With approximately 2.735M trainable critic parameters alongside an unchanged 0.45B-parameter VLA, GAF enables task outcomes to inform action generation through a compact inference-time guidance module.
\end{abstract}


\begin{minipage}{\textwidth}
\captionsetup{hypcap=false}
\centering
\includegraphics[width=0.76\textwidth]{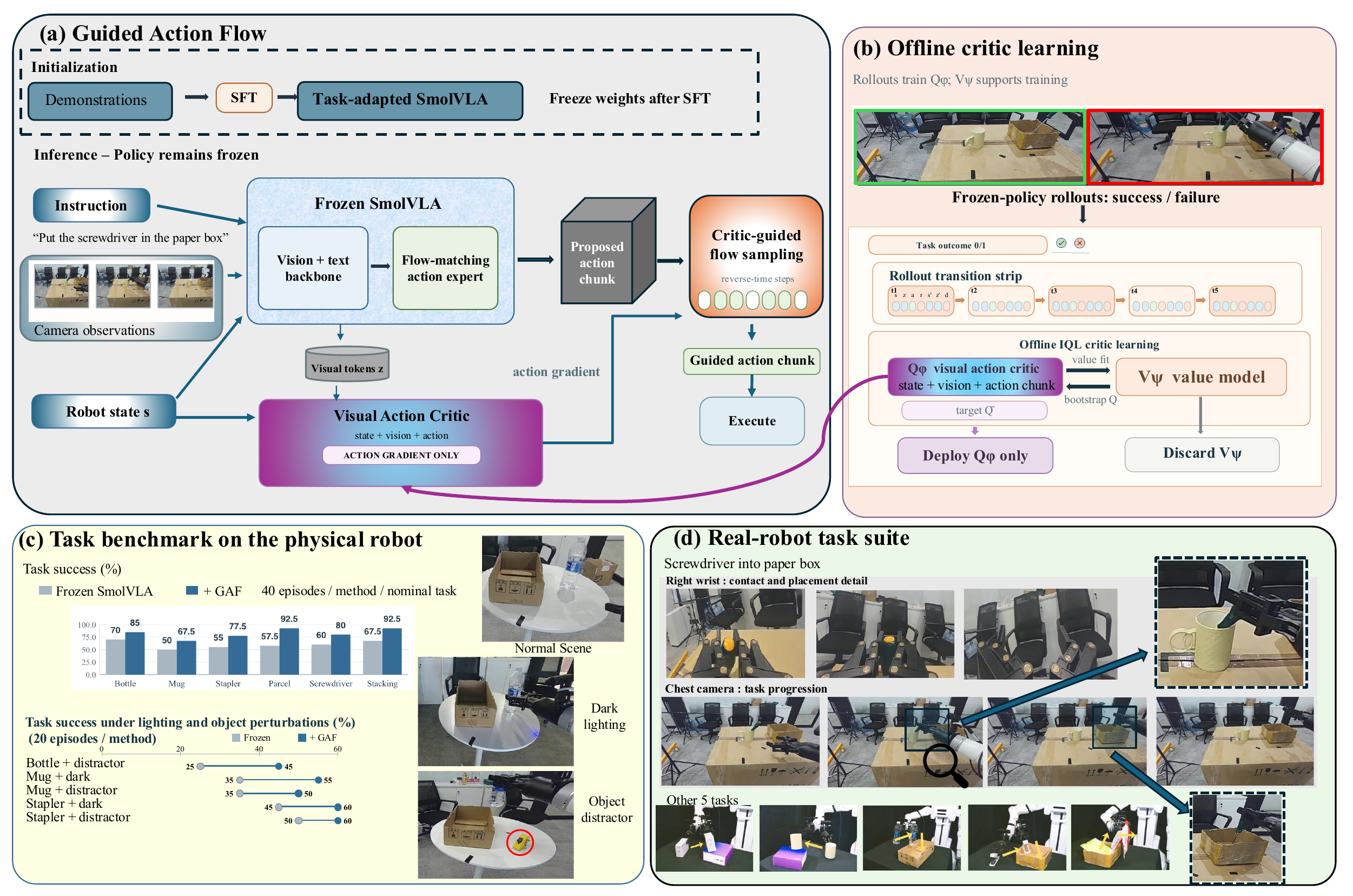}
\captionof{figure}{High-level summary of Guided Action Flow and its evaluation. (a) A compact action critic guides reverse-time sampling of a vision-language-action (VLA) policy frozen after supervised fine-tuning (SFT). (b) Task-rollout outcomes train the critic through offline implicit Q-learning (IQL). (c) Nominal task success and selected environment-perturbation results. (d) Qualitative executions from the six-task robot suite, including wrist- and chest-camera views.}
\label{fig:hero}
\end{minipage}

\section{Introduction}

Vision-language-action (VLA) models connect multimodal observations and language instructions to robot actions. Recent VLAs build on pretrained vision-language representations and robot demonstrations to support manipulation across tasks and embodiments~\cite{brohan2023rt2,kim2024openvla,black2024pi0}.

Supervised fine-tuning adapts these policies from demonstrations of desired behavior~\cite{kim2024openvla,black2024pi0}. Reinforcement-learning post-training uses task outcomes, preferences, or related behavioral signals to further optimize decisions~\cite{tan2025riptvla,li2025simplevlarl,chen2025pirl}. In these settings, feedback is incorporated during post-training; we instead ask whether task-rollout outcomes can influence action generation at deployment while the supervised-fine-tuned VLA parameters remain fixed.

Test-time guidance offers a way to apply related feedback after policy training. Existing methods use an auxiliary value, reward, or conditional objective to influence output generation without updating the policy~\cite{nakamoto2024vgps,zhou2026qgf}. A flow-matching policy forms an action chunk through successive sampling steps, providing a natural point at which such a signal can be applied.

GAF instantiates this interface for a flow-matching VLA. It learns an observation-conditioned Q critic that takes proprioception, visual features, and a candidate action chunk as input. During reverse-time sampling, the critic gradient supplies a local correction to the action-generation velocity while the supervised-fine-tuned \smolvla{} parameters remain fixed. Fig.~\ref{fig:hero} summarizes this interface and the evaluation. Guidance-direction and critic-input ablations examine the sources of the performance gains, while mechanism diagnostics characterize the critic scores and resulting action corrections. The paper therefore contributes an inference-time action-guidance layer for a frozen, supervised-fine-tuned flow-matching VLA and an evaluation across nominal tasks, controlled perturbations, ablations, and diagnostics.

\section{Related Work}

\subsection{Vision-Language-Action Policies}

Learned visual representations support object detection for manipulation~\cite{zhao2024yolomanipulation} and VLA models that map observations and instructions to actions~\cite{brohan2023rt2,kim2024openvla,tan2026signvla}. Generalist robot policies learn from demonstrations spanning tasks and embodiments, providing an initialization that can be adapted to new control settings~\cite{ghosh2024octo,kim2024openvla}. Their action representations range from discrete tokens to continuous chunks generated through diffusion or flow sampling~\cite{kim2024openvla,pertsch2025fast,ghosh2024octo,black2024pi0,shukor2025smolvla}. GAF uses the flow-matching interface as the point at which task feedback can influence inference.

\subsection{Feedback-Based Control and Policy Improvement}

Feedback-based control offers several ways to improve behavior under uncertainty, including disturbance estimation and compensation~\cite{tang2016dobc,tang2019ilc,tang2024slotcoating,tang2026disturbanceobserver} and tracking and stabilization with reinforcement learning~\cite{tang2024robusttracking,tang2024outputstabilization,tang2024optimaltracking,bai2025eventtriggeredradp}. Coordination and distributed optimization~\cite{luo2020affine,onuoha2024stress,gong2025distributedoptimization} address interactions between agents, while robustness testing~\cite{hu2025hierarchicaltesting} examines system behavior under adverse conditions. For learned policies, a key distinction is whether feedback updates the policy parameters or influences how actions are selected.

In language-model post-training, preferences, process supervision, data selection, and group-based objectives provide several routes to policy improvement~\cite{ouyang2022instructgpt,bai2022constitutional,lightman2023verify,gulcehre2023rest,shao2024deepseekmath,zheng2025gspo}; complementary analyses examine stability, efficiency, and evaluation~\cite{yang2026posttraining,yang2026rlposttrainingsurvey}.

Robot-learning systems expose related design choices under continuous actions, partial observability, and physical interaction constraints. Value-based methods estimate action values for control or policy extraction~\cite{kalashnikov2018qtopt,kostrikov2021iql}; other methods use rollout outcomes to update tokenized or continuous VLA policies directly~\cite{li2025simplevlarl,tan2025riptvla,chen2025pirl}. Offline-to-online approaches combine demonstrations, value learning, and subsequent policy updates for robot adaptation~\cite{chen2025conrft}. GAF trains an auxiliary critic from task rollouts and applies it during inference while leaving the VLA fixed.

\subsection{Test-Time Guidance}

After policy training, the same feedback can be applied at the point where a model produces its output. Test-time guidance uses an auxiliary objective to influence inference without changing the pretrained model. Classifier guidance and classifier-free guidance modify image-sampling trajectories with an additional conditional signal~\cite{dhariwal2021diffusion,ho2022classifierfree}; in sequential decision-making, learned reward or value models can similarly bias generated trajectories or action sequences~\cite{janner2022diffuser}. The common feature is that task information acts during output generation.

Candidate-based methods generate multiple proposals and use a learned evaluator to rerank or select them, as in value-guided policy steering~\cite{hansen2023idql,nakamoto2024vgps}. Gradient-based flow guidance instead modifies an intermediate sample during iterative generation and can obtain a critic gradient at an estimate of the completed action without differentiating through the full sampling trajectory~\cite{zhou2026qgf}. GAF applies this interface to a multimodal flow-matching VLA, using a task-rollout-trained critic conditioned on visual features, proprioception, and action chunks, and evaluates it on robot manipulation tasks.

\Needspace{8\baselineskip}
\section{Method}
\label{sec:method}

\subsection{System Overview}
\begin{figure}[htbp]
\centering
\includegraphics[width=0.98\textwidth]{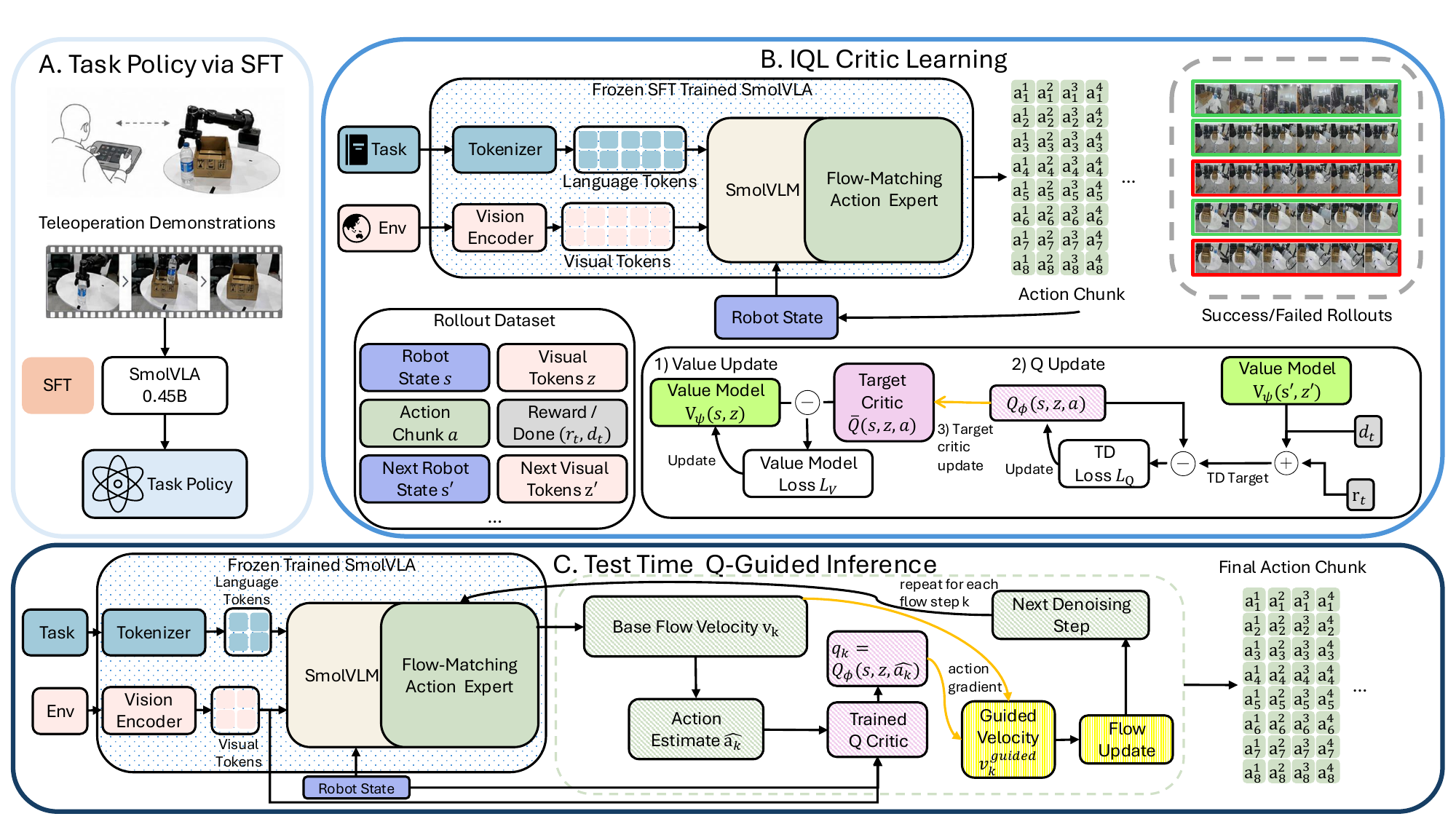}
\caption{Detailed GAF training and inference pipeline. (A) Human demonstrations are used to fine-tune SmolVLA. (B) Rollouts from the frozen policy train an action critic and a value model with offline IQL and temporal-difference (TD) critic updates. (C) At each reverse-time flow step, the critic gradient at the estimated action chunk adjusts the sampling velocity. The VLA parameters remain fixed throughout (B) and (C); the value model is used only during training.}
\label{fig:gaf_pipeline}
\end{figure}

Fig.~\ref{fig:gaf_pipeline} summarizes the GAF pipeline. Human demonstrations first initialize the task policy. The resulting \smolvla{} policy is then kept fixed while robot task rollouts provide the data for an observation-conditioned action critic. During deployment, the critic gradient modifies the reverse-time flow computation before the generated action chunk is executed. The VLA generates the base multimodal action chunk, and GAF modifies the sampling velocity used to produce it.

\subsection{Policy Initialization and Action Interface}
We first obtain the policy by supervised fine-tuning \smolvla{} on human expert demonstrations, as illustrated in Fig.~\ref{fig:gaf_pipeline}(A). We use 50 expert teleoperation demonstrations for each task, with 40 used for supervised fine-tuning and 10 held out for validation. After this initialization, all \smolvla{} parameters remain fixed during critic learning and guided inference.

At each policy step, the robot observation includes the proprioceptive state and the current camera observations. For the robot used in our experiments, the proprioceptive state is an 8-dimensional vector
\begin{equation}
    s = [q_1,\ldots,q_7,g_{\mathrm{grip}}],
\end{equation}
where $q_i$ are the seven joint angles and $g_{\mathrm{grip}}$ is the binary gripper state. The VLA maps this multimodal observation and the language instruction to a continuous action chunk
\begin{equation}
    a_{0:H-1}\in\mathbb{R}^{H\times d_a}, \qquad H=50,\quad d_a=8,
\end{equation}
containing seven joint channels and one gripper channel. Critic training and guidance both use this chunk in the policy's normalized action space.

\subsection{Task Rollout Dataset}
Following policy initialization, the frozen policy is deployed to collect task rollouts for critic training, as shown in Fig.~\ref{fig:gaf_pipeline}(B). We collect 40 rollouts per task, including successful and unsuccessful executions, to train a critic for that task. Each tuple is defined at the policy/chunk level: $s$ denotes the current proprioceptive state, $z$ the frozen vision-encoder features extracted from the current camera observations, and $a\equiv a_{0:H-1}$ the complete action chunk. After executing the chunk, $s'$ and $z'$ denote the next proprioceptive state and visual features. The binary task reward is $r\in\{0,1\}$, and $d\in\{0,1\}$ indicates whether the transition terminates the episode: nonterminal chunks have $(r,d)=(0,0)$, while the terminal chunk has $d=1$ and receives $r=1$ on success or $r=0$ on failure.

The dataset is split by episode so that chunks from one rollout do not appear in both training and validation sets. The discount factor is applied per action chunk rather than per low-level control step.

\subsection{Action Critic and IQL Training}
\begin{figure}[t]
\centering
\includegraphics[width=0.62\linewidth]{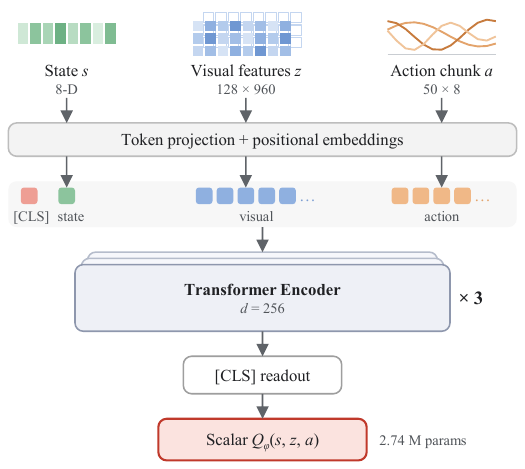}
\caption{Action-critic architecture. Proprioceptive state, visual features, and the 50-step action chunk are projected into a common token space, processed by a Transformer encoder, and mapped from the classification ([CLS]) token to the scalar action-value estimate $Q_\phi(s,z,a)$.}
\label{fig:critic_architecture}
\end{figure}

The observation-conditioned action critic $Q_\phi$ in Fig.~\ref{fig:critic_architecture} assigns a task-value estimate to an action chunk in the current observation. The critic takes the proprioceptive state, visual features, and the complete action chunk as inputs and outputs a scalar value. In the implementation, these inputs are fused by a Transformer-based critic network with a 256-dimensional hidden representation, three Transformer layers, and four attention heads. Each action critic contains approximately 2.735M trainable parameters, excluding the frozen visual feature extractor.

A separate value model $V_\psi$ is used only during offline IQL training~\cite{kostrikov2021iql}. Using the transition quantities defined above, the value model is trained by expectile regression with a slowly updated target critic $\bar{Q}$; targets are detached during each update:
\begin{equation}
    \delta = \bar{Q}(s,z,a)-V_\psi(s,z),
\end{equation}
\begin{equation}
    \mathcal{L}_V = \mathbb{E}\!\left[w(\delta)\delta^2\right],\qquad
    w(\delta)=
    \begin{cases}
        0.7,&\delta>0,\\
        0.3,&\delta\leq0.
    \end{cases}
\end{equation}
The critic is trained with the temporal-difference (TD) target
\begin{align}
    y &= r + \gamma(1-d)V_\psi(s',z'),\\
    \mathcal{L}_Q &= \mathbb{E}\!\left[(Q_\phi(s,z,a)-y)^2\right].
\end{align}
Here $d$ is the termination indicator and $\gamma$ is the chunk-level discount factor. The target critic is updated by $\bar{Q}\leftarrow\tau Q+(1-\tau)\bar{Q}$ with Polyak coefficient $\tau=0.005$. We use $\gamma=0.99$, expectile $0.7$, AdamW with learning rate $3\times10^{-4}$ and weight decay $10^{-4}$, batch size 16, and at most 80 training epochs. The checkpoint with the lowest validation TD loss is selected. The value model is discarded after training; only $Q_\phi$ is used during inference.

\subsection{GAF Guidance During Reverse-Time Flow}

The frozen \smolvla{} sampler follows a reverse-time flow convention. For the flow-matching parameterization used here, the training interpolation and target velocity are
\begin{equation}
    x_t=t\epsilon+(1-t)a, \qquad v_t^{*}=\epsilon-a,
\end{equation}
where $a$ is a clean action chunk and $\epsilon$ is Gaussian noise. At inference, the policy predicts $v_t=v_\theta(x_t,t\mid o,\ell)$, conditioned on the robot observation $o$ (images and proprioception) and language instruction $\ell$. Integration proceeds from $t=1$ to $t=0$, and the estimated clean action is
\begin{equation}
    \hat{a}=x_t-tv_t.
\end{equation}
GAF differentiates the deployed critic with respect to this estimate~\cite{zhou2026qgf},
\begin{equation}
    g=\nabla_{\hat{a}}Q_\phi(s,z,\hat{a}),
\end{equation}
clips each action-chunk gradient to a maximum $L_2$ norm of 1.0 over all $H\times d_a$ entries, and updates the velocity as
\begin{equation}
    v_t^{\mathrm{guided}}=v_t-\frac{\operatorname{clip}(g,1.0)}{\beta}.
    \label{eq:gaf}
\end{equation}
The subtraction follows from the sampler's reverse-time convention. The deployed configuration uses one critic and $\beta=2$ for every deployment condition, giving a nominal guidance coefficient of $1/\beta=0.5$ after clipping. Guidance is applied at all 10 flow steps; the uncertainty gate is disabled (equivalently, fixed to one). Gradients are computed only with respect to the action estimate and do not update \smolvla{}.

\subsection{Runtime and Safety Boundary}

The policy server receives images and robot state and returns an action chunk from the guided sampler. Existing robot-side components remain responsible for limits, motion guarding, and command execution.

\section{Experiments}

\subsection{Experimental Setup}
\begin{figure}[t]
\centering
\vspace*{2pt}
\includegraphics[width=0.70\linewidth]{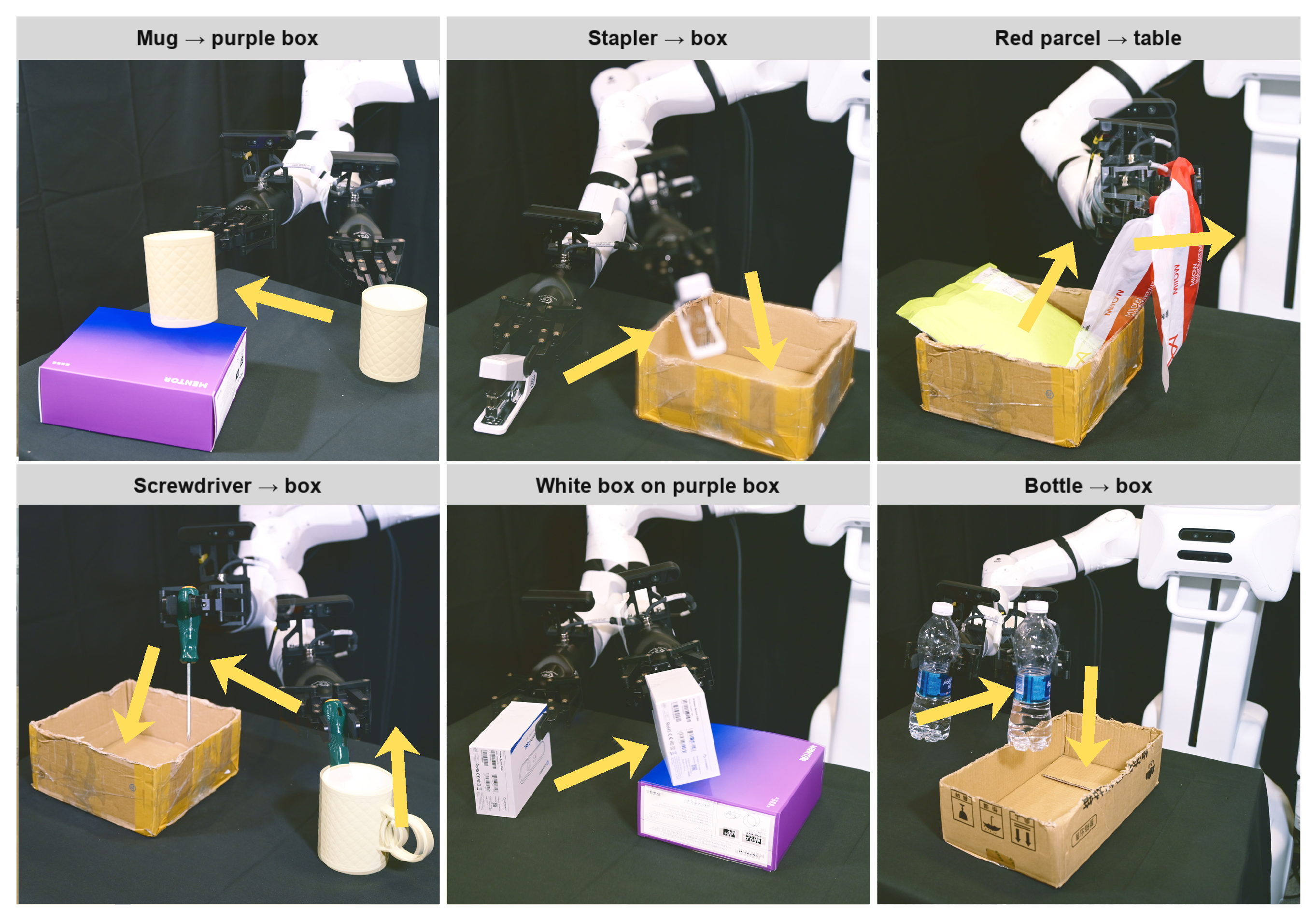}
\caption{Six-task manipulation suite. Arrows illustrate the object movements required for placement, sorting, and stacking.}
\label{fig:task_suite}
\end{figure}

We evaluate GAF on six tabletop manipulation tasks with different objects and target configurations, summarized in Fig.~\ref{fig:task_suite}. The task suite includes Bottle-to-box, Mug-to-purple-box, Stapler-into-box, Parcel sorting, Screwdriver-into-box, and Stacking boxes. All experiments use the robot's right arm, a right-wrist camera, and a chest-mounted camera. The action loop runs at 15 Hz. Each nominal policy condition is evaluated on 40 episodes. Specifically, the bottle, stapler, and screwdriver must end in an open paper box, with the screwdriver first removed from a mug; the mug and white box must rest stably on a purple box; and the red parcel must be moved from a box containing one green and one red object onto the table. We additionally evaluate three tasks under two controlled environment perturbations: altered lighting and an additional object distractor. Each perturbation policy condition uses 20 episodes.

\begin{figure}[t]
\centering
\vspace*{4pt}
\includegraphics[width=0.98\textwidth]{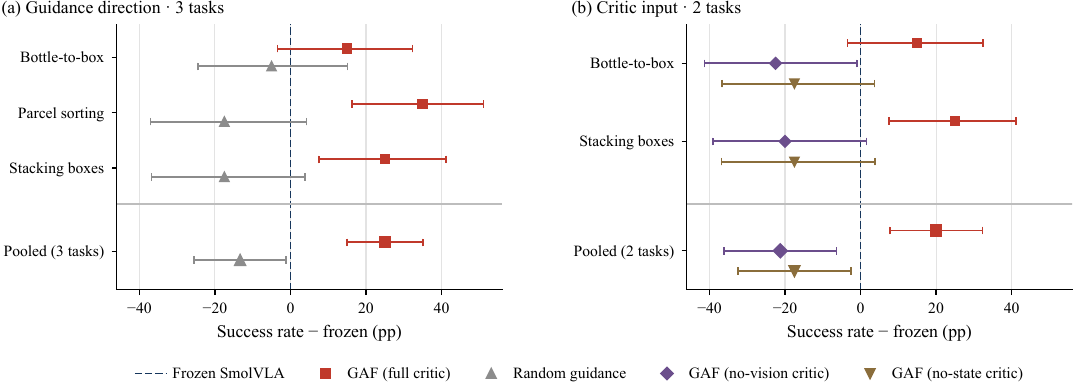}
\caption{Guidance-direction and critic-input ablations (40 episodes per method and task). Differences are relative to Frozen SmolVLA, with 95\% confidence intervals. (a) GAF with the full critic versus Random Guidance. (b) Full, no-vision, and no-state critics; VLA inputs are unchanged.}
\label{fig:ablations}
\end{figure}

An episode is successful when the manipulated object satisfies its task-specific placement, sorting, or stacking criterion. The same success definition is used for the frozen policy and GAF within each condition. Aggregate success pools episodes across the corresponding tasks or conditions. Fig.~\ref{fig:ablations} reports 95\% Newcombe intervals for individual task differences and Mantel--Haenszel intervals for pooled differences; Fig.~\ref{fig:mechanism} uses episode-bootstrap intervals.

\subsection{Nominal Task Success}
Table~\ref{tab:main_results} summarizes the nominal comparison. Across all six tasks, the frozen SmolVLA policy achieves an aggregate success rate of 60.0\%, while GAF reaches 82.5\%, an absolute improvement of 22.5 percentage points. GAF improves the observed success rate on every task in the suite.

\begin{table}[t]
\caption{Nominal task success rates (\%). $\Delta$ denotes the gain over Frozen SmolVLA in percentage points.}
\label{tab:main_results}
\centering
\footnotesize
\setlength{\tabcolsep}{4pt}
\renewcommand{\arraystretch}{1.15}
\begin{tabularx}{\linewidth}{@{}>{\raggedright\arraybackslash}p{0.38\linewidth}>{\columncolor{frozenblue}\centering\arraybackslash}X>{\columncolor{gafred}\centering\arraybackslash}Xr@{}}
\toprule
Task & Frozen & GAF & \shortstack{$\Delta$\\(pp)} \\
\midrule
Bottle & 70.0\% & \textbf{85.0\%} & \textbf{+15.0} \\
Mug & 50.0\% & \textbf{67.5\%} & \textbf{+17.5} \\
Stapler & 55.0\% & \textbf{77.5\%} & \textbf{+22.5} \\
Parcel & 57.5\% & \textbf{92.5\%} & \textbf{+35.0} \\
Screwdriver & 60.0\% & \textbf{80.0\%} & \textbf{+20.0} \\
Stacking & 67.5\% & \textbf{92.5\%} & \textbf{+25.0} \\
\midrule
Aggregate & 60.0\% & \textbf{82.5\%} & \textbf{+22.5} \\
\bottomrule
\end{tabularx}
\end{table}

\subsection{Performance under Environment Perturbations}
We next test whether the nominal improvement is retained when the task environment is changed (Table~\ref{tab:perturbation}). The perturbation suite covers altered lighting and additional object distractors on three tasks. Across the six perturbation conditions, the frozen policy achieves an aggregate success rate of 34.2\%, compared with 49.2\% with GAF, an absolute improvement of 15.0 percentage points. GAF has a positive observed difference in every tested perturbation condition.

\begin{table}[t]
\caption{Success rates (\%) under altered lighting and object-distractor conditions. $\Delta$ denotes the gain over Frozen SmolVLA in percentage points.}
\label{tab:perturbation}
\centering
\footnotesize
\setlength{\tabcolsep}{2.5pt}
\renewcommand{\arraystretch}{1.15}
\begin{tabularx}{\linewidth}{l>{\raggedright\arraybackslash}Xccc}
\toprule
Task & Condition & Frozen & GAF & \shortstack{$\Delta$\\(pp)} \\
\midrule
\rowcolor{lightingyellow} Bottle & Altered lighting & 15.0\% & \textbf{25.0\%} & +10.0 \\
\rowcolor{distractorpurple} Bottle & Object distractor & 25.0\% & \textbf{45.0\%} & +20.0 \\
\midrule
\rowcolor{lightingyellow} Mug & Altered lighting & 35.0\% & \textbf{55.0\%} & +20.0 \\
\rowcolor{distractorpurple} Mug & Object distractor & 35.0\% & \textbf{50.0\%} & +15.0 \\
\midrule
\rowcolor{lightingyellow} Stapler & Altered lighting & 45.0\% & \textbf{60.0\%} & +15.0 \\
\rowcolor{distractorpurple} Stapler & Object distractor & 50.0\% & \textbf{60.0\%} & +10.0 \\
\midrule
Aggregate &  & 34.2\% & \textbf{49.2\%} & +15.0 \\
\bottomrule
\end{tabularx}
\end{table}

\begin{figure}[t]
\centering
\vspace*{4pt}
\includegraphics[width=0.98\textwidth]{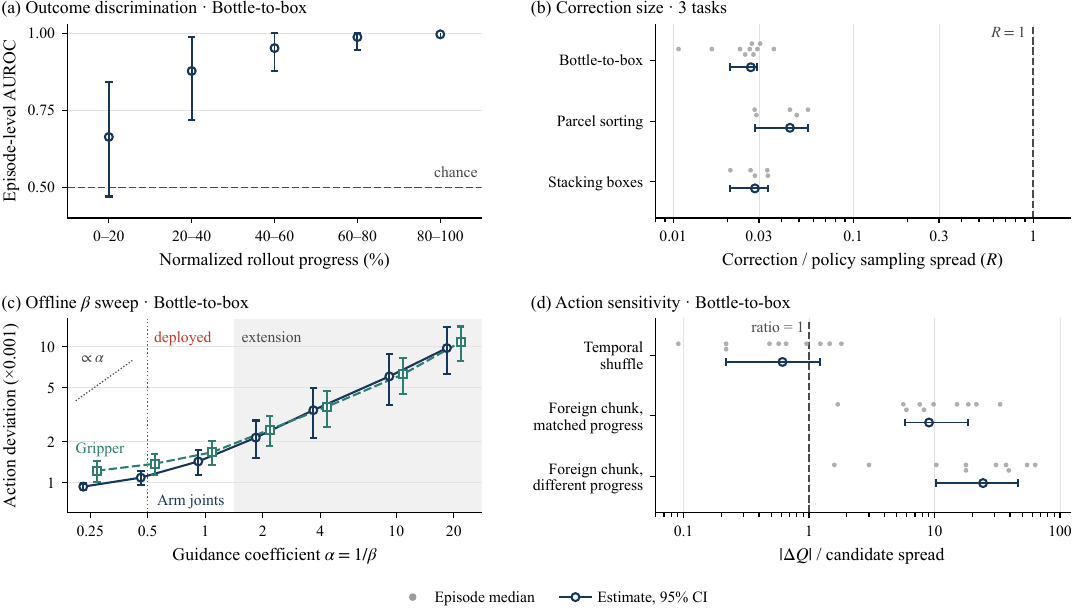}
\caption{Mechanism diagnostics for GAF. (a) Episode-level area under the receiver operating characteristic curve (AUROC) across normalized rollout progress. (b) Action correction relative to the frozen policy's action-sampling spread on three tasks, with $\beta=2$. (c) Offline Bottle-to-box sweep: root-mean-square (RMS) action deviation in normalized units versus $\alpha=1/\beta$; the deployed setting is $\alpha=0.5$. (d) Absolute critic-score change under action substitution, relative to the Q-score range of policy candidates at the same observation. Error bars show 95\% episode-bootstrap confidence intervals.}
\label{fig:mechanism}
\end{figure}

\subsection{Guidance and Critic-Input Ablations}
Fig.~\ref{fig:ablations} evaluates two properties of the guidance module. Plot (a) compares the learned critic direction with Random Guidance on Bottle-to-box, Parcel sorting, and Stacking boxes. GAF improves success by 15.0, 35.0, and 25.0 percentage points over the frozen policy, whereas Random Guidance gives differences of -5.0, -17.5, and -17.5 percentage points. Pooled across the three tasks, GAF improves success by 25.0 percentage points and Random Guidance is 13.3 percentage points below the frozen policy. The learned direction therefore provides the useful part of the intervention.

For this direction-control ablation, Random Guidance replaces only the direction of the action-gradient correction. At each guided denoising step, let $g=\nabla_{\hat a}Q_\phi(s,z,\hat a)$ be the same critic gradient used by GAF and draw an independent Gaussian tensor $u\sim\mathcal{N}(0,I)$ with the full action-chunk shape. We form
\begin{equation}
    g_{\mathrm{rand}}=\frac{\lVert g\rVert_2}{\lVert u\rVert_2+\epsilon_{\mathrm{num}}}u,
\end{equation}
where $\epsilon_{\mathrm{num}}>0$ is a small numerical stabilizer. GAF follows~\eqref{eq:gaf}, while Random Guidance uses
\begin{equation}
    v_{t,\mathrm{rand}}=v_t-\frac{\operatorname{clip}(g_{\mathrm{rand}},1.0)}{\beta}.
\end{equation}
Both variants use the same critic computation, action-chunk dimensionality, per-sample clipping, the all-10-step guidance schedule, and $1/\beta$ scaling; the uncertainty-gate path is disabled in the deployed configuration. Random Guidance therefore does not select or score candidate actions: it changes only the correction direction. Each random-guidance session uses a fresh recorded session seed and a dedicated random-number generator, so the direction sequence is reproducible conditional on that seed without changing the frozen policy's flow-noise stream.

Plot (b) compares the full critic, which receives visual features, robot state, and the action chunk, with the no-vision critic and no-state critic, which omit visual features and robot state, respectively. All three use the same VLA inputs. Pooled across Bottle-to-box and Stacking boxes, the success-rate differences relative to Frozen SmolVLA are +20.0 percentage points for the full critic, -21.3 for the no-vision critic, and -17.5 for the no-state critic. These results indicate that visual context and robot state both contribute to effective guidance.

\subsection{Mechanism Diagnostics}

Fig.~\ref{fig:mechanism} examines the learned critic and the scale of its inference-time intervention. Plot (a) uses a separate held-out cohort of 40 offline Bottle-to-box episodes, excluded from critic training and checkpoint selection. Within each progress interval, an episode is represented by its median critic score. The episode-level AUROC for separating successful and failed episodes rises from 0.664 in the first interval to 0.997 in the final interval. Taking each episode's median across intervals yields an overall AUROC of 0.970. The critic scores therefore distinguish task outcomes more clearly as execution progresses.

Plot (b) divides the norm of the GAF--baseline action difference by the median pairwise distance between frozen-policy samples at the same observation. The GAF and baseline chunks share the same initial noise. The medians across episodes are 0.027, 0.045, and 0.028 for Bottle-to-box, Parcel sorting, and Stacking boxes, respectively, indicating small action adjustments relative to the frozen policy's sampling variation.

Plot (c) shows increasing RMS action deviation as the guidance coefficient increases in an offline replay of Bottle-to-box observations. At the deployed setting $\alpha=1/\beta=0.5$, the mean deviations are $1.09\times10^{-3}$ for joint channels and $1.38\times10^{-3}$ for the gripper in normalized action units. The sweep characterizes how $\beta$ controls the size of the intervention.

\Needspace{2\baselineskip}
Plot (d) probes action sensitivity on Bottle-to-box while holding the observation fixed. We either shuffle the original chunk along its temporal dimension or replace it with a chunk from another episode at matched or different normalized progress. The absolute change in Q is normalized by the maximum-minus-minimum Q score among four frozen-policy candidate chunks generated at that observation. The medians across episodes are 0.61 for temporal shuffling, 9.04 for matched-progress substitutions, and 24.35 for different-progress substitutions. These responses show sensitivity to action content, with larger score changes for chunks drawn from other episodes, especially at different progress.

\section{Discussion}
The critic-input ablations reveal that multimodal conditioning matters beyond the VLA's action generator. Although the VLA receives the same images and robot state in all conditions, omitting either modality from the critic lowers success below the frozen baseline on the evaluated tasks. Information available to the policy therefore does not automatically compensate for missing context in the guidance signal. Together with the small correction sizes in plot (b) of Fig.~\ref{fig:mechanism}, this finding highlights the role of observation-conditioned value estimates in directing local changes to a policy proposal. Future work can extend observation-conditioned guidance across broader task and workspace variation and adapt its schedule to observed uncertainty.

\FloatBarrier
\raggedbottom
\bibliographystyle{IEEEtran}
\bibliography{references}

\end{document}